\pdfoutput=1
\documentclass[11pt]{article}
\usepackage[preprint]{acl}
\usepackage{times}
\usepackage{latexsym}
\usepackage{float}
\usepackage{subcaption}
\usepackage{float}
\usepackage[T1]{fontenc}
\usepackage[utf8]{inputenc}
\usepackage{microtype}
\usepackage{inconsolata}
\usepackage{graphicx}
\usepackage{svg}

\usepackage{amsmath} 
\usepackage{booktabs}
\usepackage{diagbox}
\usepackage{xcolor}
\usepackage{caption}
\usepackage{geometry}
\usepackage{textcomp}

\title{Assessing Human Editing Effort on LLM-Generated Texts via Compression-Based Edit Distance}

\author{
  Nicolas Devatine \\
  Tiime, Paris \\
  \texttt{nicolas.devatine@tiime.fr}\\
\And 
Louis Abraham \\
Tiime, Paris\\
  \texttt{louis.abraham@tiime.fr}
}

\begin{document}
\maketitle
\begin{abstract}
Assessing the extent of human edits on texts generated by Large Language Models (LLMs) is crucial to understanding the human-AI interactions and improving the quality of automated text generation systems. Existing edit distance metrics, such as Levenshtein, BLEU, ROUGE, and TER, often fail to accurately measure the effort required for post-editing, especially when edits involve substantial modifications, such as block operations. In this paper, we introduce a novel compression-based edit distance metric grounded in the Lempel-Ziv-77 algorithm, designed to quantify the amount of post-editing applied to LLM-generated texts. Our method leverages the properties of text compression to measure the informational difference between the original and edited texts. Through experiments on real-world human edits datasets, we demonstrate that our proposed metric is highly correlated with actual edit time and effort. We also show that LLMs exhibit an implicit understanding of editing speed, that aligns well with our metric. Furthermore, we compare our metric with existing ones, highlighting its advantages in capturing complex edits with linear computational efficiency.
Our code and data are available at: \url{https://github.com/NDV-tiime/CompressionDistance}.
\end{abstract}

\section{Introduction}

Recent advances in LLMs and text generation models have enabled the production of high-quality texts for a wide range of applications. Despite their impressive capabilities, LLMs often produce outputs that require human intervention to refine, correct, or adapt to specific contexts \citep{gehrmann2022repairing}. For example, in the context of a company using LLMs to draft customer emails, assessing the level of human intervention can guide system developers to refine models, enhance user experiences, and reduce costs by understanding how much work humans still need to perform. However, quantifying this editing effort is challenging, as it often involves not just minor fixes but also substantial restructuring or extensive content changes. Understanding the extent of human edits on LLM-generated texts is therefore essential to evaluate model performance and improve human-AI collaboration. 

Existing metrics to measure the difference between texts, such as BLEU \citep{papineni2002bleu}, ROUGE \citep{lin2004rouge}, TER \citep{snover2006study}, METEOR \citep{banerjee2005meteor}, and BERTScore \citep{zhang2019bertscore}, have been widely used in machine translation, summarization, and text generation tasks. However, these metrics often fail to capture the complexity of human edits, especially when edits involve substantial rephrasing, restructuring, or content modifications.

In the context of post-editing effort in machine translation, metrics such as HTER \citep{specia2010estimating} and CharacTER \citep{wang2016character} have been proposed to estimate the amount of human effort required to correct machine-generated translations. However, these metrics focus primarily on surface-level changes and may not fully account for the deeper semantic and structural edits commonly made on LLM-generated texts.

In this paper, we propose a novel approach for measuring human edits on LLM-generated texts using a compression-based edit distance. Our metric is inspired by the concept of compression distance, which has been explored in the context of segment rearrangements by \citet{ergun2003comparing}. Unlike traditional edit distance metrics that focus on character-level operations, such as insertions, deletions, and substitutions, \citet{ergun2003comparing} consider a richer set of operations, including segment rearrangements, like substring relocations, duplications, and deletions. Their work provides approximation algorithms for efficiently computing similarity between sequences undergoing such complex transformations. Inspired by this, our compression-based metric is designed to capture both basic edits and higher-level structural transformations, offering a more comprehensive reflection of the editing effort compared to character-level methods. Our contributions are as follows:

\begin{itemize}
    \item Efficient compression-based edit distance metric and its implementation, extending traditional edit operations to include substring-level transformations for a more comprehensive measure of human edits on LLM-generated texts.
    \item High-quality dataset of both synthetic and human edits on LLM-generated texts.
    \item Extensive set of experiments demonstrating that the proposed metric correlates highly with actual human editing time and effort, outperforming traditional metrics.
    \item Evidence from synthetic data showing that LLMs reduce the proposed distance when instructed to edit more quickly, suggesting an implicit understanding of \emph{speed} aligned with our metric.
\end{itemize}

\section{Related Work}

Measuring the similarity or difference between texts is a fundamental task in natural language processing, with applications in machine translation evaluation, text summarization, plagiarism detection, and more. Traditional edit distance metrics, such as the Levenshtein distance \cite{levenshtein1966binary}, compute the minimum number of character-level insertions, deletions, and substitutions required to transform one string into another. While useful, these metrics often fail to capture semantic differences and are sensitive to surface-level variations, like swapping two paragraphs.

In machine translation, metrics such as BLEU \cite{papineni2002bleu} and ROUGE \cite{lin2004rouge} have been widely adopted to evaluate the quality of generated translations and summaries by comparing n-gram overlaps with reference texts. However, these metrics are limited in their ability to account for paraphrasing and do not always correlate well with human judgments \cite{callison2006re}.

Translation Edit Rate (TER) \cite{snover2006study} and its human-targeted variant HTER \cite{specia2010estimating} were introduced to better measure the post-editing effort required to correct machine translations. TER considers the number of edit operations at the word level, including shifts (block movements), to transform a system output into a reference translation. However, even with these enhancements, TER may not fully capture the complexity of human edits, particularly when significant restructuring is involved.

Recent work has explored alternative approaches to better estimate the post-editing effort and capture semantic similarities. Metrics such as METEOR \cite{banerjee2005meteor} incorporate synonymy and paraphrasing through the use of linguistic resources such as WordNet \cite{miller-1994-wordnet}. BERTScore \cite{zhang2019bertscore} leverages contextual embeddings from pre-trained language models to compute similarity at the semantic level. Although these methods improve correlation with human judgments, they can be computationally intensive and may still miss structural changes.

Compression-based distances offer a different perspective by measuring the amount of information shared between sequences. The Normalized Compression Distance (NCD) \cite{cilibrasi2005clustering} is a metric derived from the Kolmogorov complexity, approximated using standard compression algorithms. NCD has been applied in various domains, including clustering and anomaly detection \cite{jiang2022low}. However, NCD can be sensitive to the choice of compression algorithm and its parameters. In the context of text classification, \citet{jiang2022low} proposed a parameter-free classification method using compressors, demonstrating that compression can be effective in low-resource settings. Their work highlights the potential of compression-based methods in deep learning and NLP tasks.

Our work focuses on a specific compression distance designed to compare sequences with segment rearrangements, as introduced by \citet{ergun2003comparing}. This distance allows operations on substrings, such as move, copy, and delete, that align closely with the types of edits humans perform when refining texts. Computing the exact segment rearrangement distance between two sequences is NP-hard; however, \citet{ergun2003comparing} proposed a constant-factor approximation using compression distances as upper bounds. The authors showed that the segment rearrangement distance can be approximated in linear time using the Lempel-Ziv-77 compression. We used the efficient algorithm proposed by \citet{crochemore2012simple} for computing the Lempel-Ziv factorization, enabling linear-time computation suitable for practical applications.

\section{Compression Distance}

Our compression-based distance metric is inspired by the work of \citet{ergun2003comparing}, who formalized sequence similarity in the presence of both character-level and segment rearrangement edits. They proposed an efficient approximation algorithm that estimates the segment rearrangement distance up to a constant factor using compression-inspired techniques.

Given two sequences \( S \) (source) and \( T \) (target), the problem involves computing the minimum number of edit operations—character edits, substring deletions, relocations, and duplications—required to transform \( S \) into \( T \). This problem generalizes the classical edit distance with additional flexibility for substring-level operations. Unfortunately, computing the exact solution is NP-hard due to the combinatorial explosion of potential edits. Thus, \citet{ergun2003comparing} proposed an approximation based on data compression, specifically leveraging properties of the Lempel-Ziv-77 (LZ77) algorithm, which provides an efficient approximation of the segment rearrangement distance. Formally, for a given sequence \( S \), LZ77 partitions \( S \) into a sequence of non-overlapping phrases\footnote{In theoretical computer science, these \textit{phrases} are often referred to as \textit{words}.} such that each new phrase is the longest match that occurs earlier in the sequence. The number of phrases in this factorization is a proxy for the complexity of \( S \). The compression distance \( d(S \to T) \) is then defined as:
\[
d(S \to T) = \text{LZ}(S \mid T) - \text{LZ}(S),
\]
where \( S \mid T \) is the concatenation of \( S \) and \( T \) with a delimiter, and \( \text{LZ}(\cdot) \) denotes the number of phrases in the LZ77 factorization. Intuitively, this measures the additional complexity of concatenating \( T \) to \( S \). \citet{ergun2003comparing} proved that it gives a 4-approximation upper bound to the exact distance, that can be efficiently computed. We apply the algorithm proposed by \citet{crochemore2012simple} for computing the Lempel-Ziv factorization in linear time from suffix arrays. as implemented in \texttt{pydivsufsort}\footnote{Which itself relies on \url{https://github.com/y-256/libdivsufsort} for suffix array computation.} \cite{pydivsufsort}.

\section{Experimental Settings}
\label{sec:experiments}

We now turn to the empirical evaluation of our compression-based edit distance. Our goals are to: (1) validate that it correlates strongly with actual human editing effort, and (2) compare it against traditional and well-established edit distance metrics.

\subsection{Datasets}
\label{subsec:datasets}

In the following, we introduce a new dataset of LLM-generated answers to accounting questions. This dataset includes both \textit{synthetic edits} produced by an LLM itself (under different editing scenarios) and \textit{human edits} produced by expert annotators. Additionally, we experimented with the publicly available IWSLT 2019 post-editing dataset \citep{scarton-etal-2019-estimating}.

\begin{figure}[htbp!]
    \centering
    \includegraphics[width=1.0\linewidth]{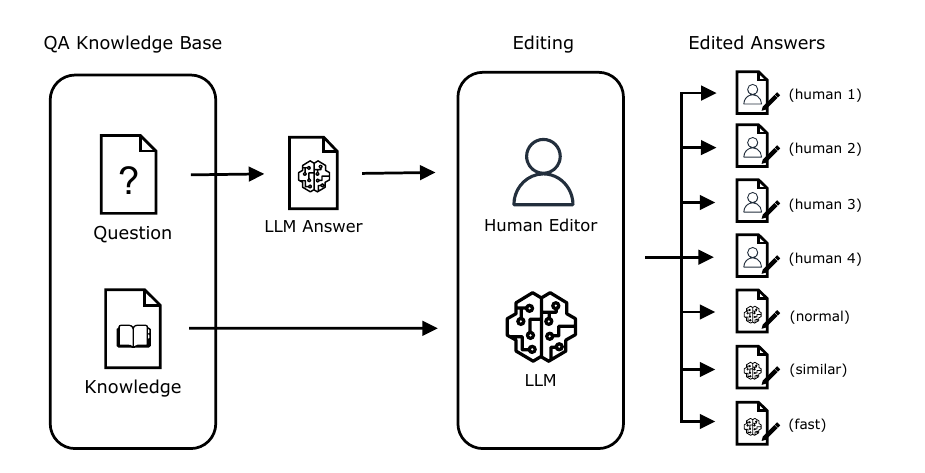}
    \caption{Overview of our dataset construction process. We sampled $200$ questions (and associated \emph{expert knowledge}) from a Q\&A knowledge base. First, each question is answered by an LLM without being provided the \textit{expert knowledge}. Then, these answers are edited by either a human or an LLM with the \textit{expert knowledge} provided, resulting in a final post-edited answer. Three scenarios are considered when editing is done by an LLM: \emph{normal}, \emph{similar}, \emph{fast}). For human edits, we measured the edit times (in seconds).}
\end{figure}

\begin{figure*}[h!]
    \centering
    \includegraphics[width=\textwidth]{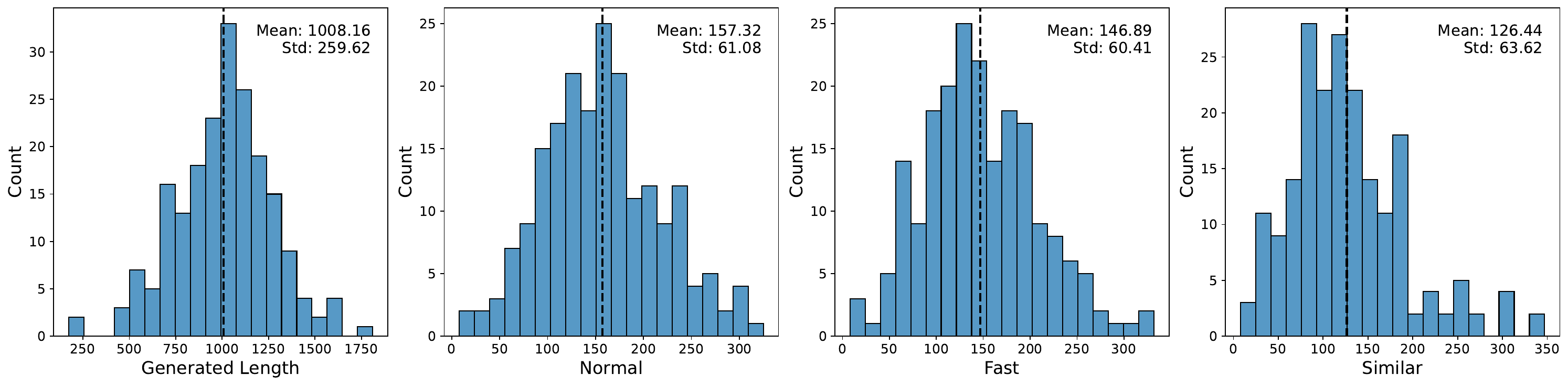}
    \caption{Length distribution of initial LLM answers and distribution of the compression-based edit distances for the different editing scenarios in the synthetic dataset.}
    \label{fig:histograms}
\end{figure*}

\subsubsection{Accounting Q\&A Edits}

We first randomly sampled 200 French accounting-related questions from a proprietary Q\&A knowledge base. Each question has an associated expert gold standard answer that we refer to as \textit{expert knowledge} throughout this paper, providing accurate domain-specific information. We then used \texttt{claude-3-sonnet-20240229} (March 2024) from Anthropic \citep{anthropic2024claude3} to generate initial answers to these questions \emph{without} providing the LLM with the \emph{expert knowledge}, potentially leading to wrong or incomplete answers.

\paragraph{Human Edits.}
We recruited four expert annotators with domain experience in accounting. Each annotator received the same set of 200 initial LLM-generated responses, together with the relevant associated \emph{expert knowledge}. They were asked to correct errors, update content, and generally improve the answers. Each expert-produced edit was recorded together with its editing time (in seconds), measured from the moment the annotator received the initial answer to the submission of the edited version. This yielded a total of $4 \times 200 = 800$ human-edited answers, each with an associated edit time. Since these expert edits reflect genuine human post-editing effort, they provide a valuable benchmark for validating the correlation of our proposed metric with real editing behavior.

\paragraph{Synthetic Edits.} 
In addition to human edits, we created a synthetic dataset of edits by prompting a new instance of the same LLM to revise the initial answers after being provided with the relevant \emph{expert knowledge}. The translated version of the prompts used are provided in Appendix \ref{sec:appendix_prompts}. Specifically, for each answer, we asked the LLM to perform edits in three distinct scenarios: (i) \emph{normal edits}, where the LLM is instructed to correct and improve the answer using the newly provided domain knowledge; (ii) \emph{similar edits}, where the LLM is instructed to preserve the main structure of the initial answer but to incorporate the new expert content to correct and improve the answer; and (iii) \emph{fast edits}, where the LLM is prompted to edit the initial answers as quickly as possible, simulating minimal revision effort. This process resulted in three synthetic edits per question, yielding a total of $3 \times 200 = 600$ synthetic post-edited answers. Each synthetic edit is grounded in the same reference \emph{expert knowledge} but follows different editing constraints, allowing us to evaluate the sensitivity of the metric to various types and depths of revisions.

\subsubsection{IWSLT 2019}
For further comparison with existing edit distance metrics, we draw on a publicly available dataset \cite{scarton-etal-2019-estimating}\footnote{\url{https://github.com/carolscarton/iwslt2019}}. This dataset consists of 1{,}047 English--Spanish segments (totaling 26{,}875 words) that were machine-translated by 41 different systems. Five professional translators performed a post-editing task on every segment. The translators used the PET tool \citep{aziz-etal-2012-pet}, which logs all edit operations along with the time taken to post-edit. However, block edit operations are not directly supported in this tool. The dataset thus provides a rich set of human edits, including the total edit time and the number of keystrokes per edit.

\subsection{Evaluation}

\begin{table}[!htbp]
\centering
\small
\begin{tabular}{@{}l|cc@{}}
\toprule
\textbf{Metric} & \textbf{W/o Knowledge} & \textbf{W/ Knowledge} \\ \midrule
Compression Dist. & \textbf{0.71} & \textbf{0.81} \\
Levenshtein & 0.68 & 0.59 \\
CharacTER & 0.62 & 0.53 \\
TER & 0.42 & 0.28 \\
BERTScore & -0.63 & -0.56 \\
ROUGE-L & -0.62 & -0.62 \\
BLEU & -0.62 & -0.47 \\
METEOR & -0.59 & -0.37 \\ \bottomrule
\end{tabular}
\caption{Pearson correlation between metrics and human edit times. Correlations are measured with (\textit{W/ Knowledge}) and without (\textit{W/o Knowledge}) concatenating the expert knowledge text to the initial LLM output.}
\label{tab:pearson_correlations}
\end{table}

\begin{table*}[!h]
\centering
\begin{tabular}{@{}l|ccccc||ccccc@{}}
\toprule
& \multicolumn{5}{c||}{\textbf{Keystrokes}} & \multicolumn{5}{c}{\textbf{Edit Time}} \\
\textbf{Metric} & \textbf{A0} & \textbf{A1} & \textbf{A2} & \textbf{A3} & \textbf{A4} & \textbf{A0} & \textbf{A1} & \textbf{A2} & \textbf{A3} & \textbf{A4} \\ \midrule
Compression Dist. & \textbf{0.87} & 0.85 & \textbf{0.82} & \textbf{0.89} & \textbf{0.86} & \textbf{0.65} & \textbf{0.59} & 0.70 & \textbf{0.73} & 0.55 \\
Levenshtein. & 0.85 & \textbf{0.91} & \textbf{0.82} & \textbf{0.89} & 0.83 & 0.64 & 0.55 & \textbf{0.72} & \textbf{0.73} & \textbf{0.56} \\
TER & 0.52 & 0.35 & 0.54 & 0.56 & 0.56 & 0.28 & 0.26 & 0.36 & 0.38 & 0.28 \\
BLEU & -0.50 & -0.42 & -0.51 & -0.51 & -0.51 & -0.29 & -0.31 & -0.36 & -0.37 & -0.28 \\
METEOR & -0.57 & -0.60 & -0.65 & -0.64 & -0.65 & -0.27 & -0.25 & -0.34 & -0.34 & -0.27 \\
ROUGE-L & -0.52 & -0.59 & -0.51 & -0.52 & -0.52 & -0.28 & -0.36 & -0.33 & -0.32 & -0.26 \\
BERTSc. & -0.56 & -0.64 & -0.56 & -0.58 & -0.58 & -0.32 & -0.41 & -0.40 & -0.41 & -0.32 \\
CharacTER & 0.51 & 0.64 & 0.55 & 0.53 & 0.54 & 0.28 & 0.35 & 0.39 & 0.35 & 0.30 \\ \bottomrule
\end{tabular}
\caption{Pearson correlations of various metrics with keystrokes and edit time on the IWSLT2019 dataset. A0--A4 refer to the five different annotators in that dataset.}
\label{tab:pearson_combined}
\end{table*}

We evaluated how well our compression-based edit distance and various baselines correlate with human post-editing effort across the different datasets. First, we computed the distance values for all tested metrics between the initial text and its edited version. We use Pearson's correlation as our primary statistic, as it provides a straightforward measure of linear relationship between a metric’s output and actual editing effort. All reported correlation coefficients are statistically significant ($p < 0.05$). We compare our compression-based edit distance with the following standard metrics: BLEU \citep{papineni2002bleu}, ROUGE-L \citep{lin2004rouge}, TER \citep{snover2006study} and HTER \citep{specia2010estimating}, Levenshtein Distance, METEOR \citep{banerjee2005meteor}, CharacTER \citep{wang2016character}, and BERTScore \citep{zhang2019bertscore}.

\begin{figure}[h!]
    \centering
    \includegraphics[width=\columnwidth]{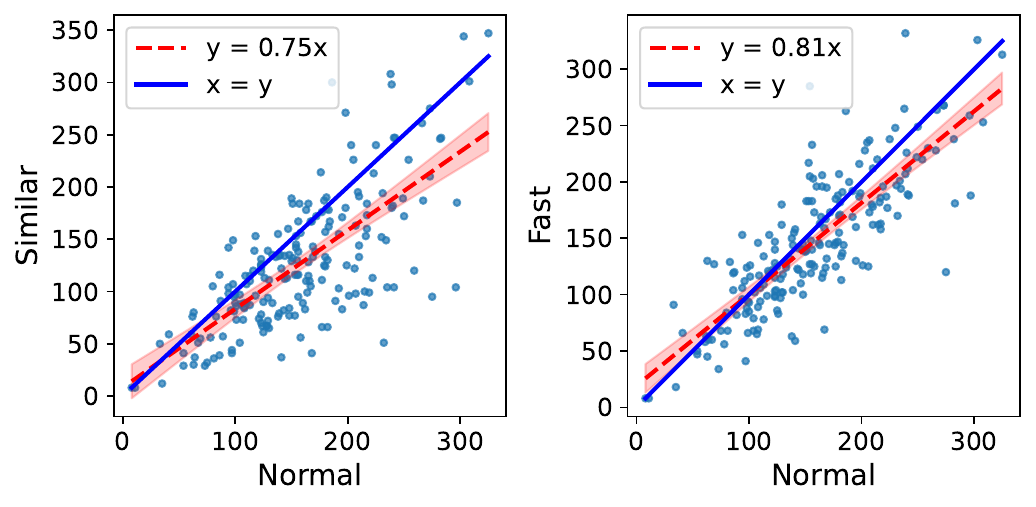}
    \caption{Comparison of compression distances on our synthetic dataset between the \emph{normal} editing scenario and the \emph{similar} (left) and \emph{fast} (right) editing scenarios. Each subplot shows scatter points and a fitted linear regression line (in red). The blue line $x=y$ is shown for reference.}
    \label{fig:regression}
\end{figure}

\paragraph{Pearson Correlation with Edit Time.}
For our newly introduced dataset of human edits, we computed the Pearson correlation between each distance metric and the edit times. To account for possible copy-paste usage from the \emph{expert knowledge} text, we compare correlations under two conditions: \emph{(i)} the metric computed using only the initial LLM output, and \emph{(ii)} the metric computed after concatenating the \emph{expert knowledge} text to the initial LLM output. In the second scenario, partial reuse of knowledge text would lower the distance for metrics that detect substring-level repetition (e.g., our compression-based edit distance). To further examine how each metric aligns with human effort, we visualize \emph{edit time vs.\ distance metric} using scatter plots coupled with simple linear regressions (see Figure~\ref{fig:metric_comparisons}). For the IWSLT2019 dataset, we compute the Pearson correlations between all metrics and \emph{(i)} edit time, and \emph{(ii)} total keystrokes performed.

\paragraph{KNN Regression.} For both the human post-edit dataset and IWSLT2019, we additionally trained a simple K-Nearest Neighbors (KNN) regressor ($K=5$) for each metric, individually predicting edit time (or keystrokes) from single-metric features. For IWSLT2019, we trained a single model by gathering the data from all annotators.  As suggested by \citet{jiang2022low}, simple classification or regression setups using compression metrics can be quite effective. We randomly split each dataset into 80\% train and 20\% test. Tables~\ref{tab:r2_individual} and~\ref{tab:r2_results} report the $R^2$ on the test sets.

\paragraph{Synthetic Edits.}
Finally, we examine the synthetic scenarios described (\emph{normal}, \emph{similar}, and \emph{fast} edits). We compare compression distance across these synthetic edits, plotting pairwise regressions to visualize whether the compression distance differentiates adequately between the different editing depths/styles. We also provide distributions of the generated text lengths and compression distances (Figure~\ref{fig:histograms}).

\section{Results}

\begin{figure*}[htbp!]
    \centering
    \begin{subfigure}{0.32\textwidth}
        \includegraphics[width=\textwidth]{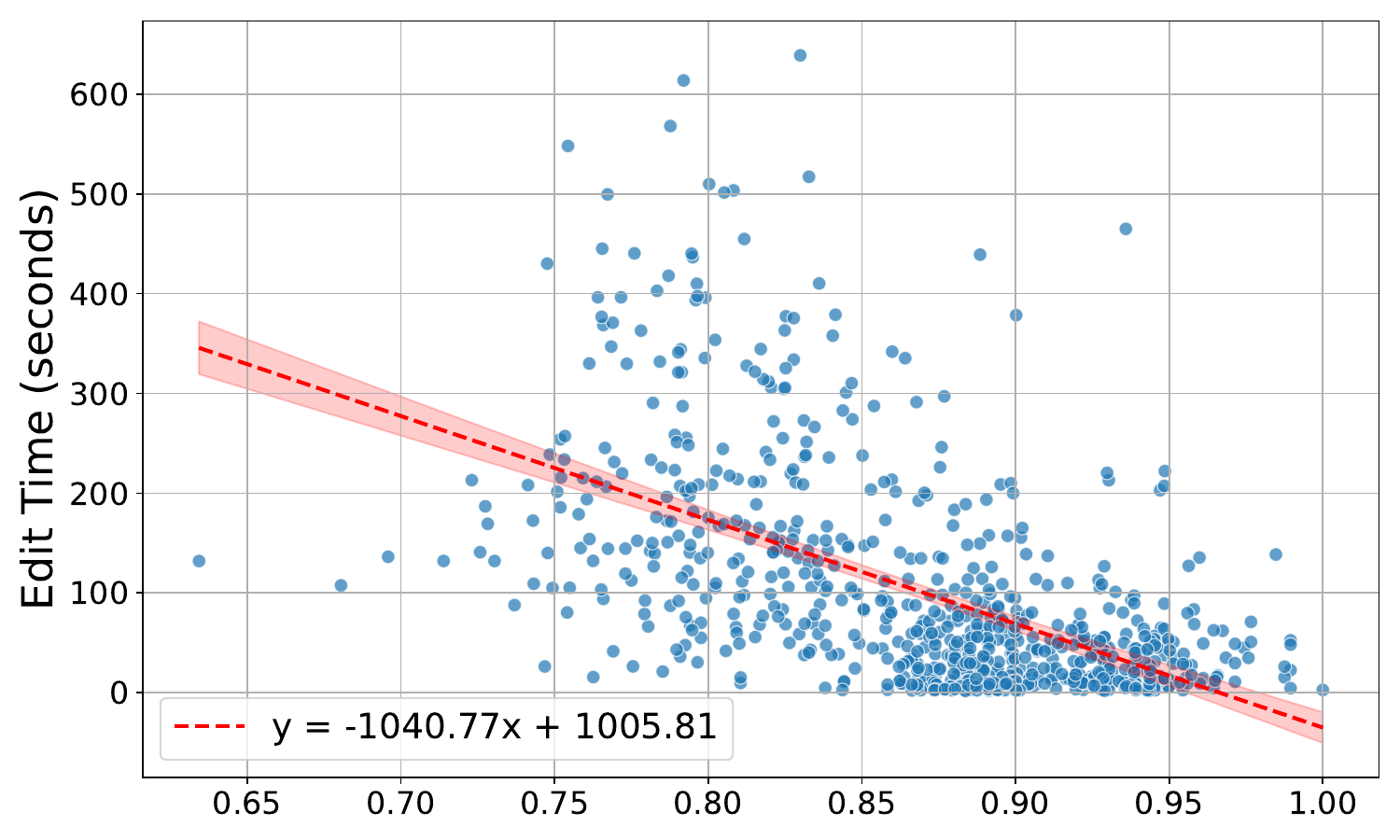}
        \caption{BERTScore vs. Edit Time}
        \label{fig:bertscore}
    \end{subfigure}
    \hfill
    \begin{subfigure}{0.32\textwidth}
        \includegraphics[width=\textwidth]{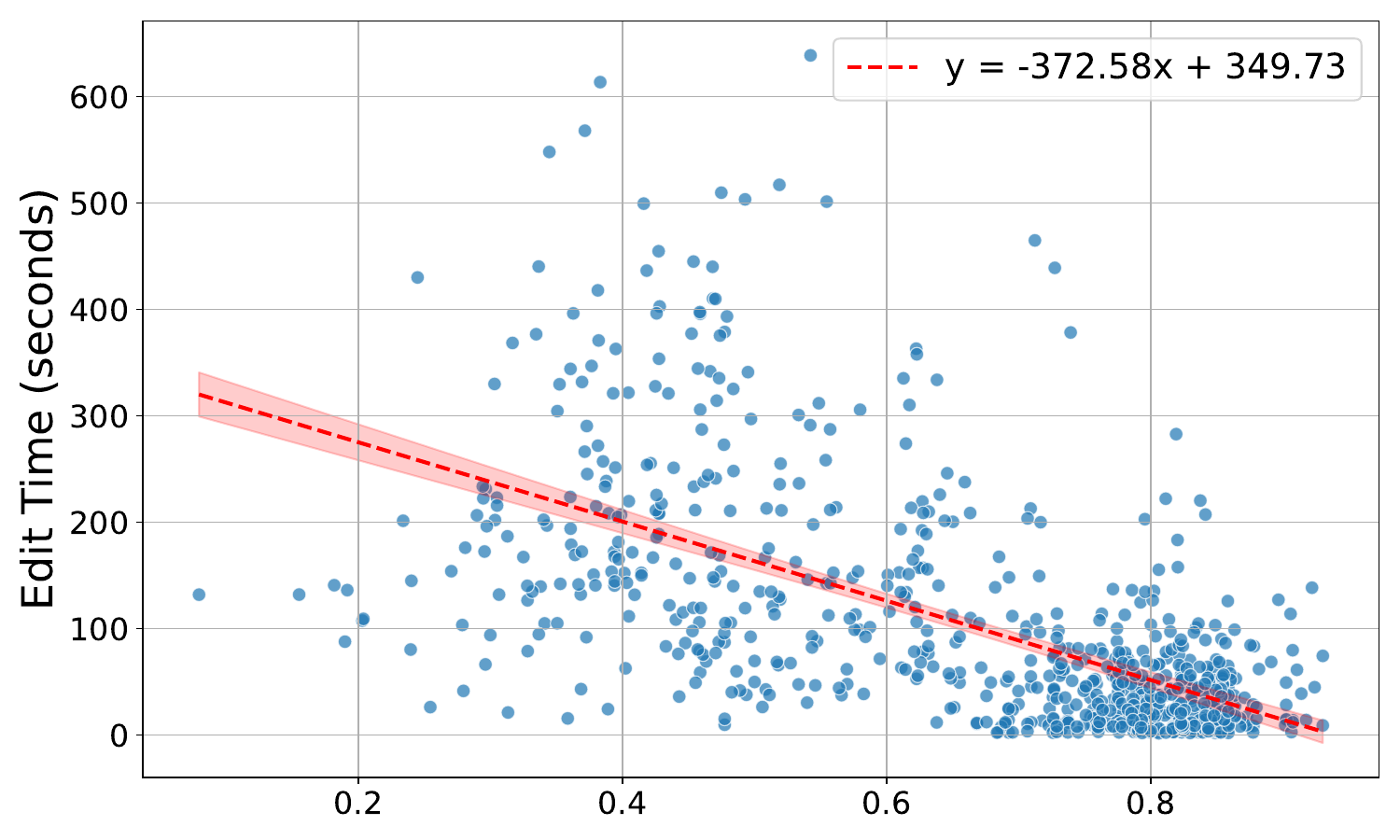}
        \caption{ROUGE-L vs. Edit Time}
        \label{fig:rougel}
    \end{subfigure}
    \hfill
    \begin{subfigure}{0.32\textwidth}
        \includegraphics[width=\textwidth]{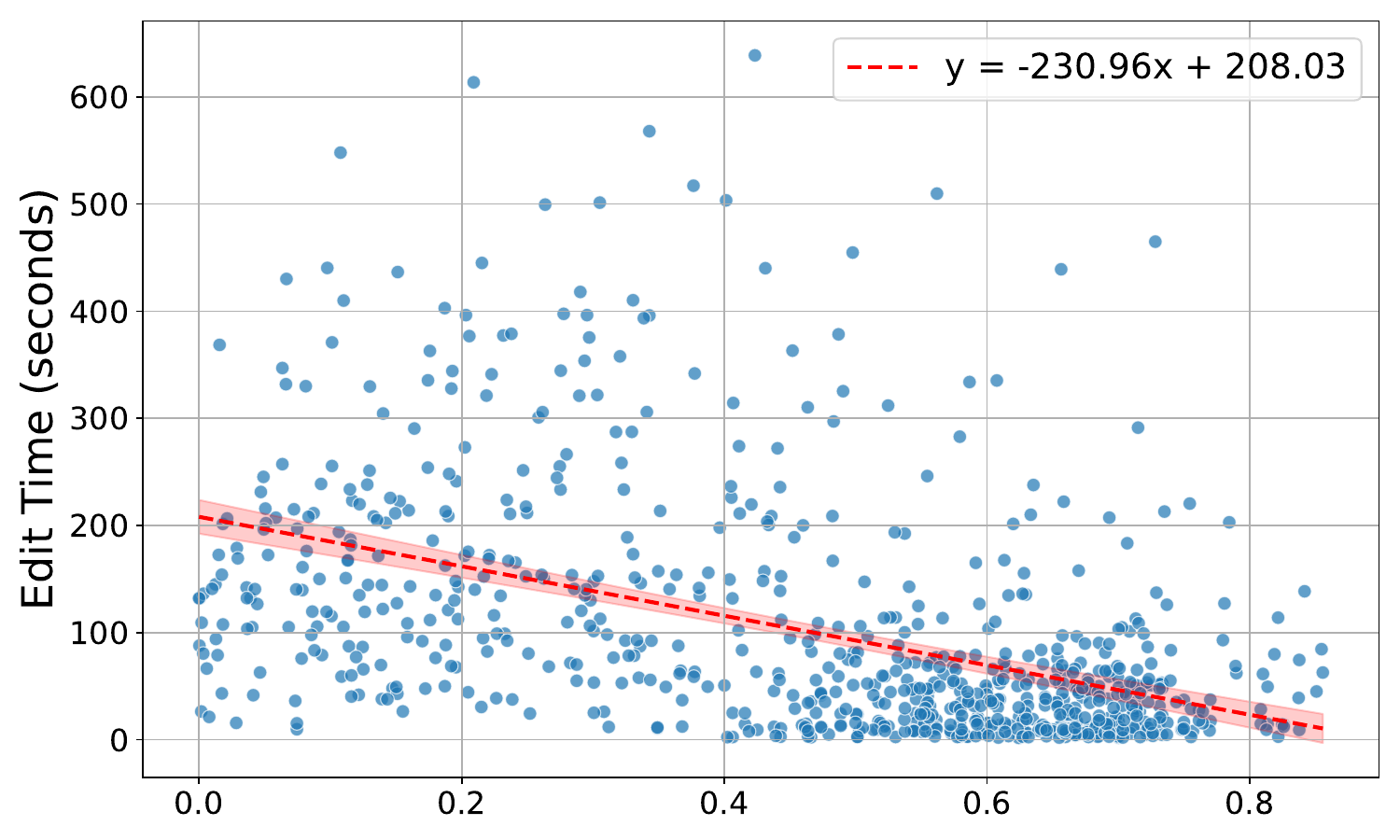}
        \caption{BLEU vs. Edit Time}
        \label{fig:bleu}
    \end{subfigure}
    
    \begin{subfigure}{0.32\textwidth}
        \includegraphics[width=\textwidth]{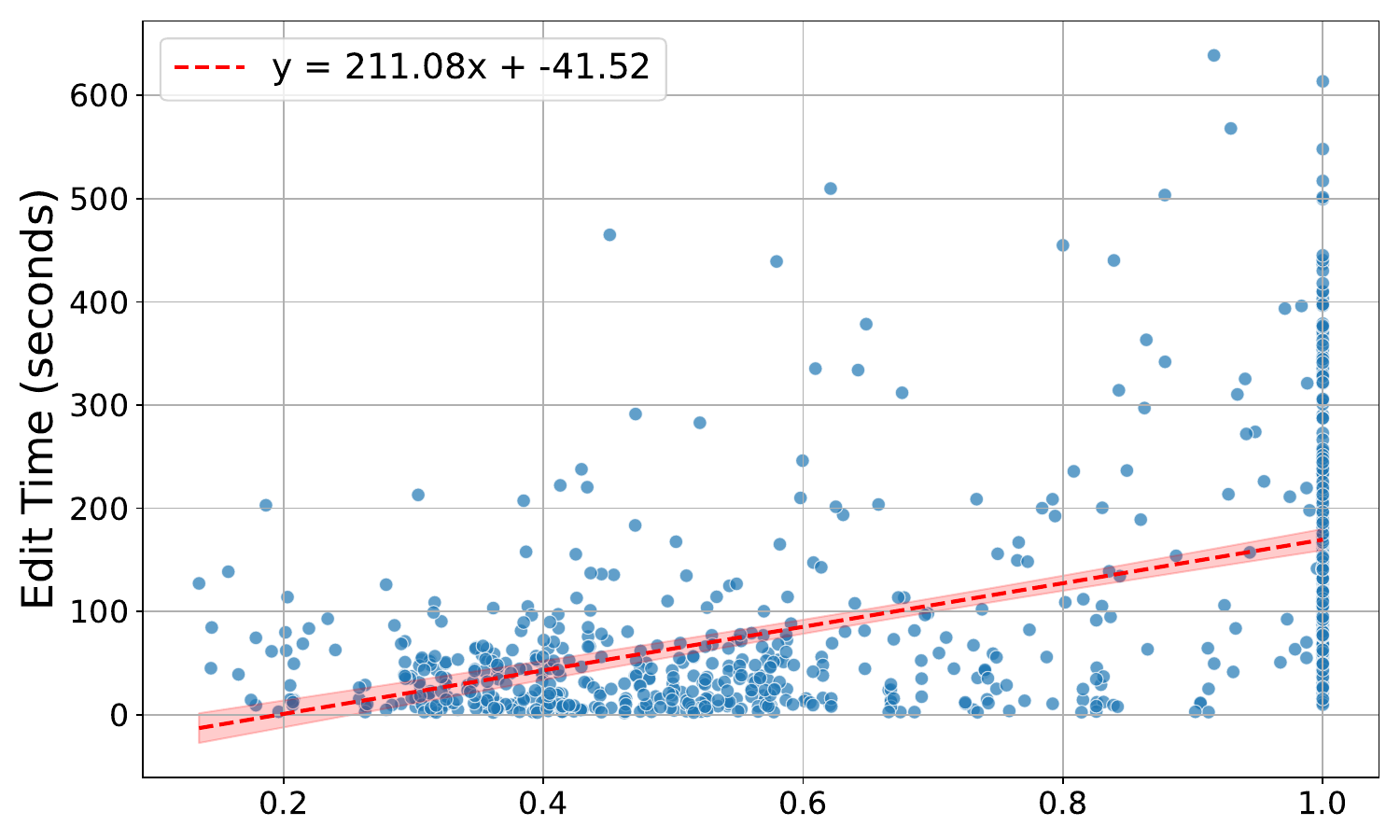}
        \caption{CharacTER vs. Edit Time}
        \label{fig:charcter}
    \end{subfigure}
    \hfill
    \begin{subfigure}{0.32\textwidth}
        \includegraphics[width=\textwidth]{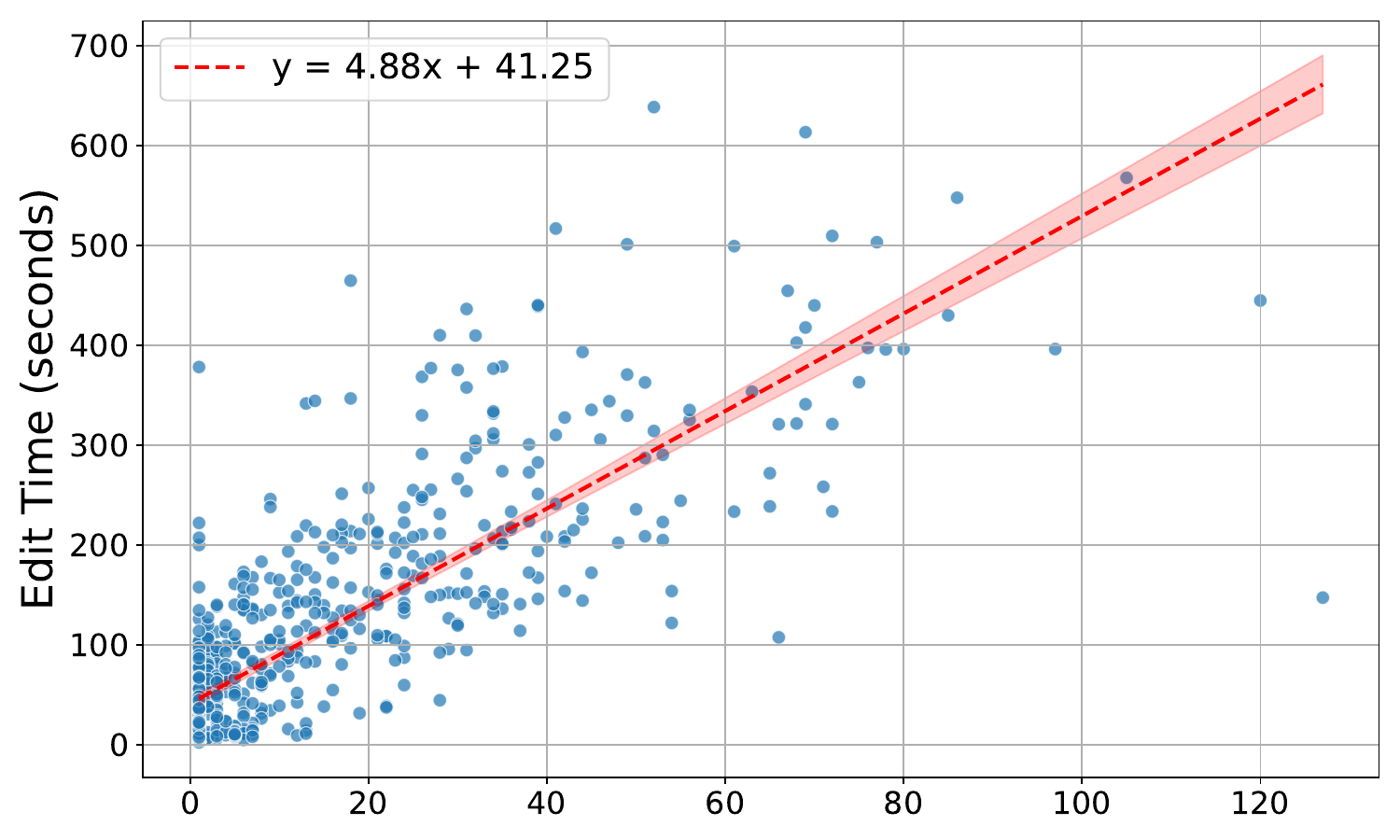}
        \caption{Compression Distance vs. Edit Time}
        \label{fig:compression}
    \end{subfigure}
    \hfill
    \begin{subfigure}{0.32\textwidth}
        \includegraphics[width=\textwidth]{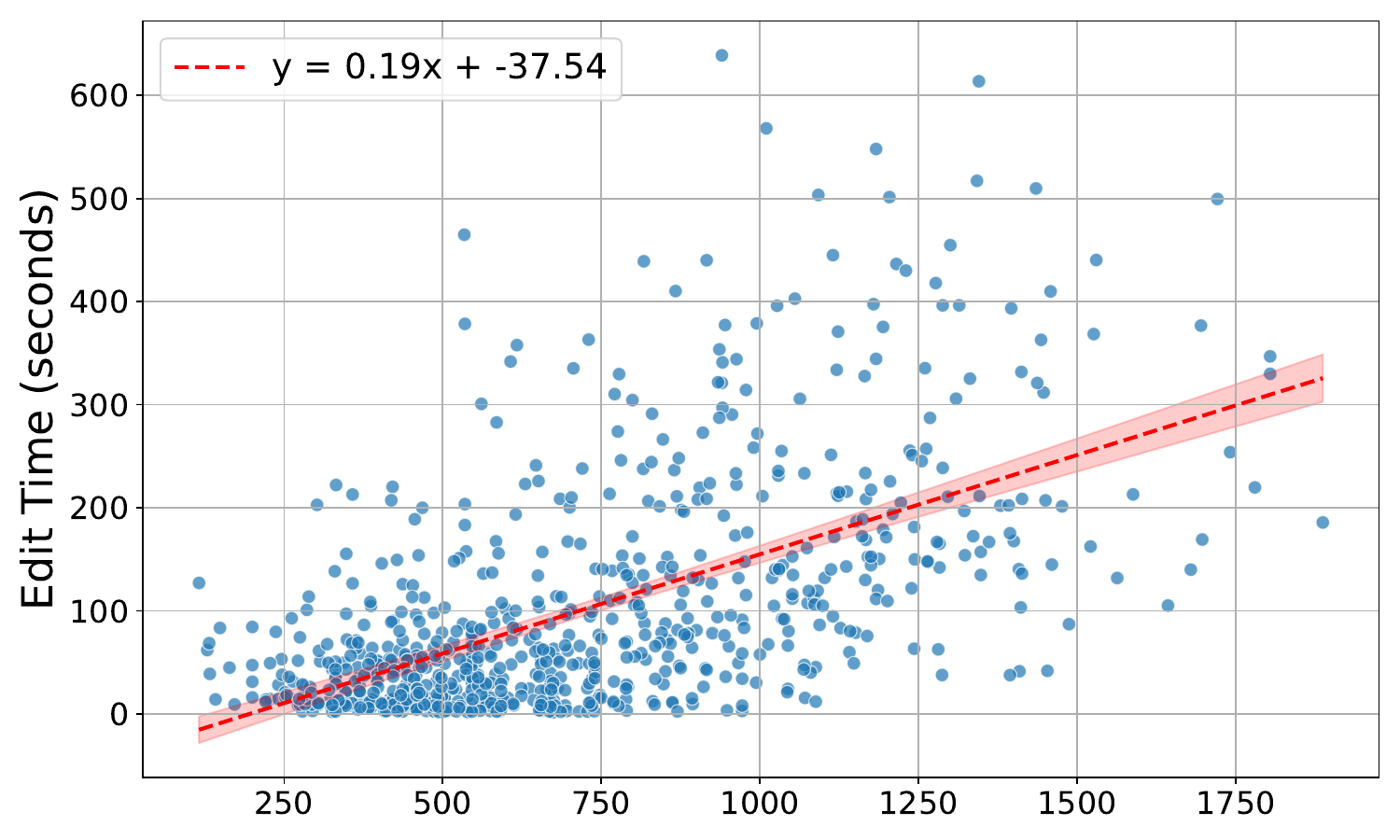}
        \caption{Levenshtein vs. Edit Time}
        \label{fig:levenshtein}
    \end{subfigure}
    
    \begin{subfigure}{0.32\textwidth}
        \includegraphics[width=\textwidth]{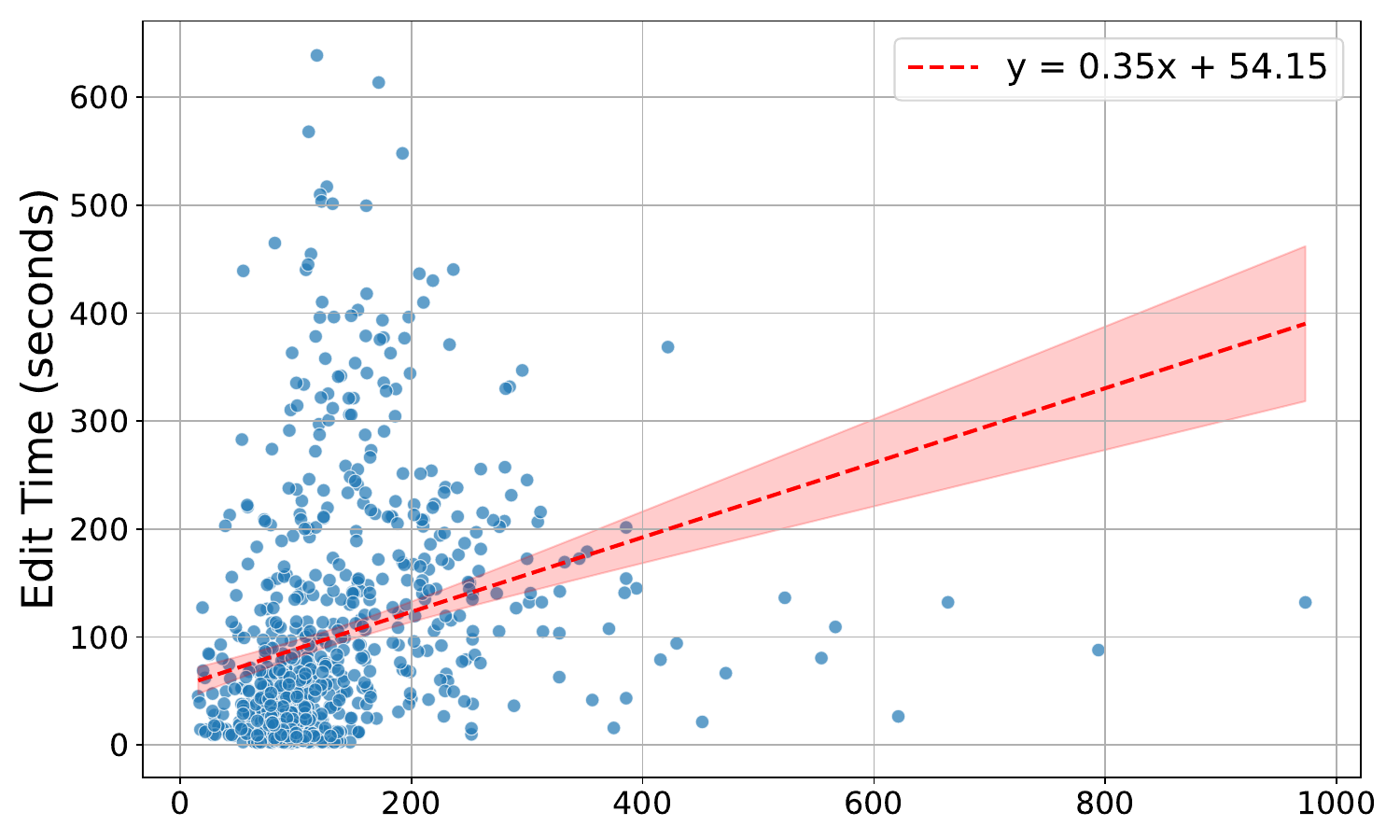}
        \caption{TER vs. Edit Time}
        \label{fig:ter}
    \end{subfigure}
    \hfill
    \begin{subfigure}{0.32\textwidth}
        \includegraphics[width=\textwidth]{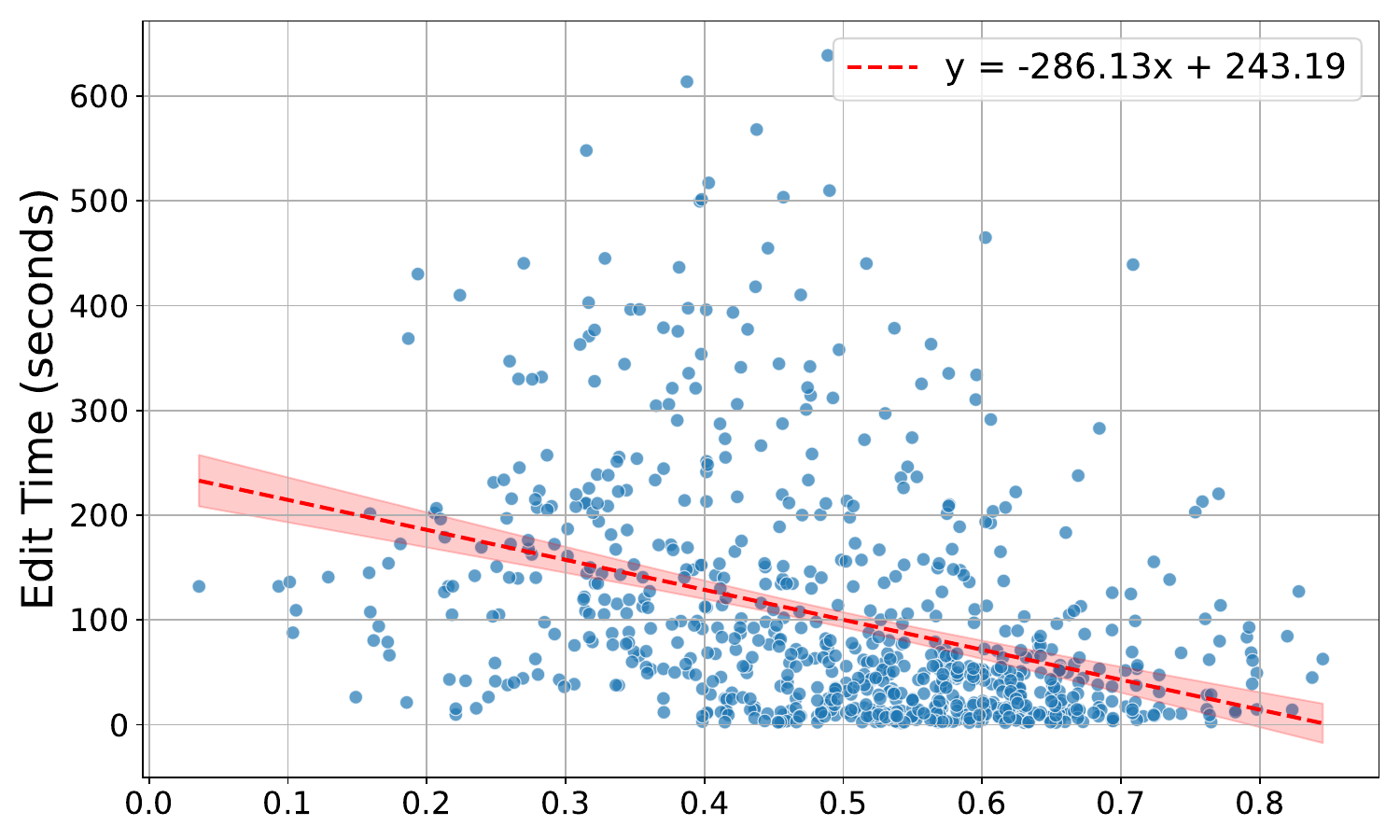}
        \caption{METEOR vs. Edit Time}
        \label{fig:meteor}
    \end{subfigure}
    
    \caption{Scatter plots with linear regression fits and confidence intervals for various metrics against measured edit times on our human post-edited dataset when concatenating the \emph{expert knowledge} text to the initial LLM output.}
    \label{fig:metric_comparisons}
\end{figure*}

From Table~\ref{tab:pearson_correlations}, our compression-based distance consistently shows stronger correlations with human editing times than most other metrics, notably reaching $0.81$ when the original LLM output is concatenated with \emph{expert knowledge}. The Levenshtein distance also exhibits a relatively high correlation ($0.59$--$0.68$), but tends to be outperformed by the compression distance when the knowledge text is included. BLEU, ROUGE-L, and BERTScore exhibit generally weaker correlations with observed edit times, suggesting that $n$-gram or semantic overlap alone is less effective at capturing complex structural edits. Overall, these findings suggest that when relying on externally provided knowledge for corrections (e.g., copy-pasting relevant text), metrics that model substring-level transformations (such as our compression-based method) better reflect actual editing time. Figure~\ref{fig:metric_comparisons} visually compares the edit distance metrics against measured edit times on our human edits using linear regressions, again revealing that traditional metrics do not effectively capture the edit time. In contrast, our compression-based metric aligns the best with actual edit times.

Table~\ref{tab:pearson_combined} confirms the consistency of compression distance on the IWSLT2019 dataset, showing robust correlations with both edit time and keystrokes across multiple annotators. Although the Levenshtein distance demonstrates comparable results, this could be attributed to the fact that block operations, which the compression distance is particularly effective at measuring (as shown in Table~\ref{tab:pearson_correlations}), are not directly supported by the PET edit tool. Other metrics such as TER or CharacTER lag behind in capturing post-editing effort.

Tables~\ref{tab:r2_individual} and~\ref{tab:r2_results} present the $R^2$ scores from our 5-NN regressor approach. On the IWSLT2019 dataset (Table~\ref{tab:r2_individual}), the compression distance yields a high $R^2$ for keystrokes (0.6053) and a moderate $R^2$ for the edit time, closely matching Levenshtein. On our human post-edited dataset (Table~\ref{tab:r2_results}), the compression distance emerges as the top metric for predicting actual edit times, confirming its ability to model the cognitive and manual aspects of editing when used in the context of LLM-generated texts. We also tested a range of metric combinations, but these multi-feature regressors did not provide meaningful improvements in $R^2$.

\begin{table}[h!]
\centering
\small
\begin{tabular}{@{}l|cc@{}}
\toprule
\textbf{Metric} & \textbf{R$^2$ (Time)} & \textbf{R$^2$ (Keys)} \\
\midrule
Compression Dist. & 0.2209 & \textbf{0.6053} \\
Lev. Dist. & \textbf{0.2879} & 0.5814 \\
BERTScore & 0.0208 & 0.2096 \\
ROUGE-L & 0.0920 & 0.3176 \\
CharacTER & -0.0052 & 0.1912 \\
METEOR & 0.0243 & 0.1873 \\
BLEU & 0.0358 & 0.0302 \\
TER & 0.1080 & 0.3462 \\
\bottomrule
\end{tabular}
\caption{KNN regression results (R$^2$) for $K = 5$ on the IWSLT2019 dataset, using a single model for all annotators, with single-metric features to predict edit time and keystrokes.}
\label{tab:r2_individual}
\end{table}

\begin{table}[h!]
\centering
\small
\begin{tabular}{@{}l|c@{}}
\toprule
\textbf{Metric} & \textbf{R$^2$ (Time)} \\
\midrule
Compression Dist.     & \textbf{0.6316} \\
Lev. Dist.      & 0.3332 \\
BERTScore       & 0.2622 \\
ROUGE-L         & 0.2609 \\
CharacTER       & 0.1246 \\
METEOR          & 0.1209 \\
BLEU            & 0.2642 \\
TER             & 0.0334 \\
\bottomrule
\end{tabular}
\caption{KNN regression results (R$^2$) for $K = 5$ with single-metric features to predict edit time on our human-edited dataset.}
\label{tab:r2_results}
\end{table}

Regarding our synthetic edits, Figure~\ref{fig:histograms} and \ref{fig:regression} show that the compression distances in the \emph{similar} scenario are typically smaller than those in the \emph{normal} and \emph{fast} scenarios. A linear regression between \emph{normal}-edit compression distances and each of the other two scenarios reveals slopes around $0.75$--$0.81$. This reflects that the faster or more structure-preserving the edit, the lower the compression distance, again suggesting the metric tracks editing intensity in a manner consistent with expectations. When instructed to edit quickly, LLMs understand and respond to the concept of editing speed and time spent, as a human would do.

\section{Discussion}
\label{sec:discussion}

Our experimental results indicate that substring-level transformations, as captured by compression-based distances, mirror the true effort involved in post-editing. Linear correlation patterns confirm that these substring-based metrics better predict editing times and keystrokes compared to traditional $n$-gram overlap measures or purely semantic similarity scores.

A notable outcome is the improved correlation of compression distance when the \emph{expert knowledge} text is concatenated to the original LLM output. This suggests that editors frequently copy or reuse sentences directly from the \emph{expert knowledge} to refine the answers. Traditional edit distances (e.g., Levenshtein) are not as robust to such copy-paste patterns, yielding weaker correlations once the knowledge text is included. In contrast, our compression-based metric detects repeated substrings of arbitrary length, effectively capturing the reuse of entire segments from the \emph{expert knowledge text}. Although the Levenshtein distance also showed high correlations, it tends to over-penalize large blocks of reordering or insertion. It is particularly apparent when block edit operations are allowed (which was not the case in the IWSLT2019 dataset).

Our additional regression experiments offer complementary insights. Regression with single features provides a coherent measure of how strongly each metric alone can account for editing variance. In particular, the compression distance outperforms other metrics in predicting both the edit time (Tables~\ref{tab:r2_results} and \ref{tab:r2_individual}) and keystrokes. Interestingly, attempts to combine multiple metrics within KNN did not yield further gains. One possible explanation is that the compression distance already capture a large portion of the editing signal, making the additional information from $n$-gram overlaps or semantic similarity somewhat redundant.

Our synthetic edits reveal that in the scenario in which the LLM is asked to produce \emph{fast} edits, it results in smaller compression distances than in the \emph{normal} edit scenario (Figure~\ref{fig:regression}), indicating reduced editing effort. This result highlights that the LLM's understanding of \emph{fast} editing, when explicitly instructed, is aligned with a reduced edit distance and, notably, with the compression distance. Furthermore, the regression plot and distance distributions reveal that the compression distances of edits produced when the LLM is instructed to preserve the main structure of the initial answer or to edit quickly (the \emph{similar} and \emph{fast} scenarios) are lower compared to unconstrained edits. This indicates that the LLM demonstrates a human-like understanding of varying levels of editing effort.

Another important aspect is the computation time. Many of the commonly used metrics (e.g., BLEU, ROUGE, and Levenshtein distance) have quadratic (or higher) complexities that are acceptable for moderately sized texts. Our compression-based approach can be performed in linear time relative to text length \citep{ergun2003comparing,crochemore2012simple}, making it scalable for large documents compared to resource-intensive metrics (e.g., BERTScore).

\section{Conclusion}

We introduced a novel compression-based edit distance metric that leverages substring-level transformations to better capture the complexity of human editing effort on LLM-generated texts. Our experiments, conducted on real-world datasets, demonstrated that this metric is highly correlated with actual human editing times and efforts, outperforming traditional metrics. By providing a more accurate reflection of both cognitive and mechanical effort in human post-editing of LLM outputs, our metric enables precise and efficient evaluation of language model performance in text generation tasks.


\bibliography{custom}

\appendix

\section{Prompts}
\label{sec:appendix_prompts}

This appendix details the prompts used in our experiments to generate and edit answers. Although the original prompts were in French to match our dataset's language, we provide English translations here. The original French prompts are provided in our code repository.

\subsection{Initial Answer Generation}
The first prompt was designed to generate baseline answers without specialized knowledge:

\begin{quote}\itshape
"You have to answer an accounting question asked in an email. I'm going to give you the question <QUESTION>. You must answer the question in detail and precisely. Answer with only the content of the answer and nothing else."
\end{quote}

\subsection{Editing Scenarios}
For the editing phase, we developed three distinct prompts corresponding to our experimental scenarios:

\paragraph{Normal Edit Scenario}
This prompt allows for unrestricted editing while incorporating \emph{expert knowledge}:

\begin{quote}\itshape
"You have to answer an accounting question asked in an email. I am going to give you the question <QUESTION>, an answer generated by a non-specialized model <LLM\_ANSWER>, and specific and specialized knowledge that allows you to answer this question <KNOWLEDGE>. You must edit the model answer based on the knowledge passed on in order to improve the answer where necessary. You can modify the answer as much as you like in order to improve the initial answer with this knowledge. You should not copy and paste the knowledge into the answer, but use the relevant elements of this knowledge to update the initial answer. Answer with only the content of the answer and nothing else."
\end{quote}

\paragraph{Similar Edit Scenario}
This prompt emphasizes preserving the original structure while improving content:

\begin{quote}\itshape
"You have to answer an accounting question asked in an email. I am going to give you the question <QUESTION>, an answer generated by a non-specialized model <LLM\_ANSWER>, and specific and specialized knowledge that allows you to answer this question <KNOWLEDGE>. You must edit the model answer based on the knowledge passed on in order to improve the answer where necessary. You must modify the answer in a way that keeps the same structure and outline as the initial answer, by modifying if necessary only the content based on the knowledge transmitted. You should not copy and paste the knowledge into the answer, but use the relevant elements of this knowledge to update the initial answer. Answer with only the content of the answer and nothing else."
\end{quote}

\paragraph{Fast Edit Scenario}
This prompt prioritizes quick, efficient edits:

\begin{quote}\itshape
"You have to answer an accounting question asked in an email. I am going to give you the question <QUESTION>, an answer generated by a non-specialized model <LLM\_ANSWER>, and specific and specialized knowledge that allows you to answer this question <KNOWLEDGE>. You must edit the model's answer based on the knowledge passed on in order to improve the answer where necessary. You must take as little time as possible to edit the initial answer while completing the task correctly. You should not copy and paste the knowledge into the answer, but use the relevant elements of this knowledge to update the initial answer. Answer with only the content of the answer and nothing else."
\end{quote}

Each prompt was designed to maintain consistency in the basic task structure while introducing specific constraints or objectives that characterize the different editing scenarios analyzed in our study.

\end{document}